\documentclass[lettersize,journal]{IEEEtran}
\usepackage{amsmath,amsfonts}
\usepackage{algorithmic}
\usepackage{algorithm}
\usepackage{array}
\usepackage[caption=false,font=normalsize,labelfont=sf,textfont=sf]{subfig}
\usepackage{textcomp}
\usepackage{stfloats}
\usepackage{url}
\usepackage{verbatim}
\usepackage{graphicx}
\usepackage{cite}
\usepackage{soul}
\usepackage{color}
\usepackage{multirow}
\usepackage{utfsym}
\usepackage{graphicx}
\usepackage{amsmath}
\usepackage{amssymb}
\usepackage{booktabs}
\usepackage{colortbl}
\usepackage{fontawesome}
\usepackage{mathtools}
\usepackage{arydshln} 
\usepackage{hyperref}
\usepackage{threeparttable}
\begin{document}

\title{DQEN: Dual Query Enhancement Network\\ for DETR-based HOI Detection}

\author{Zhehao Li, Chong Wang \textsuperscript{\faEnvelope}, Yi Chen, Yinghao Lu, Jiangbo Qian, Jiong Wang, and Jiafei Wu
\thanks{Manuscript received 28 February 2025; This work was supported by the Ningbo Municipal Natural Science Foundation of China (No. 2022J114), National Natural Science Foundation of China (No. 62271274), Ningbo S\&T Project (No.2024Z004) and Ningbo Major Research and Development Plan Project (No.2023Z225).}
\thanks{Zhehao Li, Chong Wang, Yi Chen, Yinghao Lu, Jiangbo Qian, and Jiong Wang are with the Faculty of Electrical Engineering and Computer Science, Ningbo University, Ningbo, Zhejiang 315211, China, E-mail: lllzzzhhh1019@163.com, wangchong@nbu.edu.cn.}
\thanks{Jiafei Wu is with Zhejiang Lab, Hangzhou, China. E-mail: {wujiafei@zhejianglab.com}.}
\thanks{\faEnvelope \ Corresponding Author: Chong Wang.}}

\markboth{Journal of \LaTeX\ Class Files,~Vol.~14, No.~8, August~2021}%
{Shell \MakeLowercase{\textit{et al.}}: A Sample Article Using IEEEtran.cls for IEEE Journals}

\IEEEpubid{0000--0000/00\$00.00~\copyright~2021 IEEE}


\maketitle
\begin{abstract}
Human-Object Interaction (HOI) detection focuses on localizing human-object pairs and recognizing their interactions. Recently, the DETR-based framework has been widely adopted in HOI detection. In DETR-based HOI models, queries with clear meaning are crucial for accurately detecting HOIs. However, prior works have typically relied on randomly initialized queries, leading to vague representations that limit the model's effectiveness. Meanwhile, humans in the HOI categories are fixed, while objects and their interactions are variable. Therefore, we propose a Dual Query Enhancement Network (DQEN) to enhance object and interaction queries. Specifically, object queries are enhanced with object-aware encoder features, enabling the model to focus more effectively on humans interacting with objects in an object-aware way. On the other hand, we design a novel Interaction Semantic Fusion module to exploit the HOI candidates that are promoted by the CLIP model. Semantic features are extracted to enhance the initialization of interaction queries, thereby improving the model's ability to understand interactions. Furthermore, we introduce an Auxiliary Prediction Unit aimed at improving the representation of interaction features. Our proposed method achieves competitive performance on both the HICO-Det and the V-COCO datasets. The source code is available at \url{https://github.com/lzzhhh1019/DQEN}.
\end{abstract}


\section{Introduction}
 
\label{sec:intro}
\IEEEPARstart{H}{uman-Object} Interaction (HOI) detection is an interesting task in the field of visual recognition. The goal is to localize human-object pairs and recognize their interactions, serving as a basis for advanced vision tasks that require deeper image understanding and reasoning, including image captioning ~\cite{herdade2019image,li2023cascade}, action recognition ~\cite{wang2021multi,fan2021understanding,su2023transductive}, video comprehension \cite{wang2023exploring,tao2024feature,liu2024evcap}, and visual question answering \cite{chen2020counterfactual}.

\begin{figure}[htbp]
\centering
\includegraphics[width=\columnwidth]{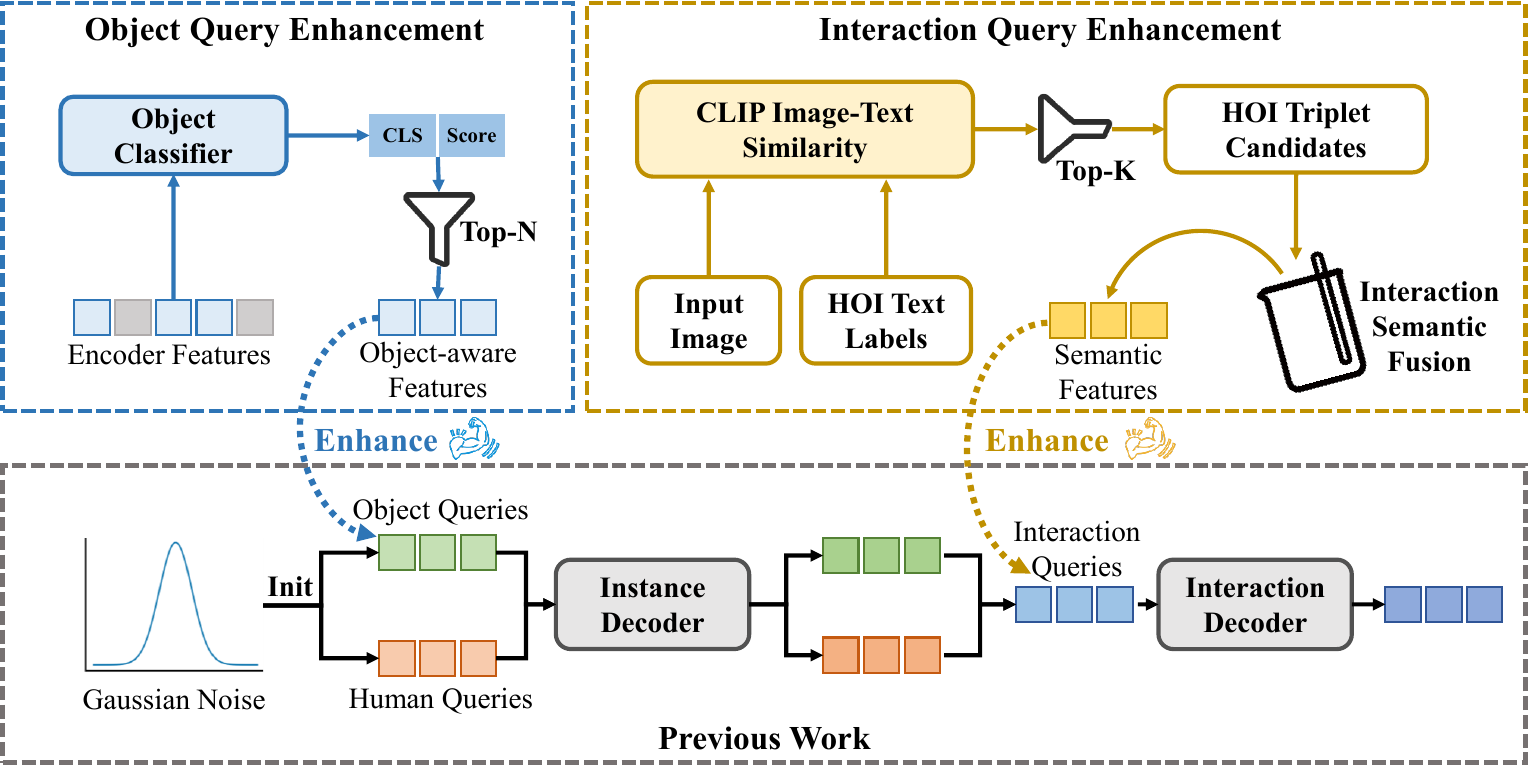}
\caption{The pipeline of proposed Dual Query Enhancement Network. Previous frameworks initialized human and object queries with Gaussian Noise and utilized the instance decoder output directly as interaction queries. Our model leverages object-aware features and semantic features to enhance object and interaction queries, respectively.
}
\label{fig:fig1}
\end{figure}

Recently, thanks to the remarkable performance of the DEtection TRansformer (DETR) \cite{carion2020end}, there has been growing interest in using this framework for HOI detection models. DETR treats object detection as an end-to-end set prediction task and introduces a set-based loss via bipartite matching. This approach can be adapted to HOI detection by utilizing a set of learned human and object queries. Subsequently, by analyzing the overall scene and interactions between humans and objects, a set of HOI triplets in the format $<$\textit{human, verb, object}$>$ can be predicted, providing a more contextualized understanding of the image content.

In the standard DETR approach, decoder queries are not image-specific and are initialized with Gaussian Noise in a stochastic manner. As illustrated within the gray dashed box in Fig. \ref{fig:fig1}, this initialization method is also widely applied in numerous prior DETR-based HOI models \cite{zhang2021mining,liao2022gen,ning2023hoiclip,tamura2021qpic}. Whereas previous research \cite{zhu2020deformable,yao2021efficient,zhang2022dino} on DETR by selecting Top-\textit{K} encoder features as priors to enhance decoder queries. In DETR-based object detection models, such an approach has been proven to improve detection accuracy. Motivated by its success, we aim to transfer this strategy to HOI detection. To the best of our knowledge, our method is the first to extend this approach to the HOI domain. Given that we can derive prior knowledge from the inherent structure of DETR, it is reasonable to ask whether we can also acquire additional prior knowledge from external sources.

\IEEEpubidadjcol

With the success of Contrastive Language-Image Pre-Training (CLIP) \cite{radford2021learning} in various downstream tasks, such as open-vocabulary and zero-shot learning problems, some one-stage HOI models \cite{liao2022gen,ning2023hoiclip} attempt to transfer CLIP's prior knowledge to HOI detection, aiming to enhance model performance. Though these methods have significantly improved performance, they primarily focus on explicitly leveraging CLIP's existing knowledge, i.e. directly utilizing CLIP's text or image features, and making minor adjustments to query settings. However, these approaches overlook the inherent image-text matching capabilities of CLIP and the importance of query initialization. Based on this perspective and motivated by CATN \cite{dong2022category}, a transformer-based HOI model that initializes object queries with category-aware semantic features, we intend to leverage CLIP’s image-text matching capability to incorporate semantic features, which contribute to enhancing the interaction queries.

Building upon the above analysis, we present the Dual Query Enhancement Network, which consists of two enhancement modules for object and interaction queries. Concretely, for the Object Query Enhancement (OQE, the blue dashed box in Fig. \ref{fig:fig1}), we select the Top-\textit{N} encoder features according to the object classifier's class scores to enrich the object queries. However, experimental results reveal that it does not enhance object detection accuracy in HOI detection. Interestingly, instead of guiding the model’s focus toward objects, it shifts attention more toward the humans interacting with the objects, which surprisingly leads to improved HOI detection performance. For the Interaction Query Enhancement (IQE, the yellow dashed box in Fig. \ref{fig:fig1}), we employ CLIP to compute the image-text similarity between the input image and the HOI Text Labels. This allows us to retrieve high-quality HOI Triplet Candidates based on the Top-\textit{K} similarity scores. An Interaction Semantic Fusion module is designed to extract the semantic features from the HOI candidates, which are then provided to the interaction queries to enhance their contextual understanding. Finally, we introduce an Auxiliary Prediction Unit to further boost the model's ability to predict $<$\textit{verb}$>$, integrating with the interaction prediction, which strengthens the model's performance in recognizing interactions.

The contributions of our work are three-fold,
\begin{itemize}
\item The initialization of interaction queries is greatly enhanced by incorporating interaction semantic features, which are extracted by a sophisticated interaction semantic fusion module from a set of HOI triplet candidates. In this procedure, the image-text matching capability of CLIP models is exploited to generate coarse guesses for HOI prediction.
\item The initialization of object queries is also enhanced by introducing object-aware encoder features. It helps the model pay more attention to the humans interacting with those objects, which contributes to improved HOI detection performance.
\item Additional $<$\textit{verb}$>$ prediction is inserted by reusing the interaction semantic features, which further improves the accuracy of recognizing interactions.

\end{itemize}

\begin{figure*}[htbp]
\centering
\includegraphics[width=\textwidth]{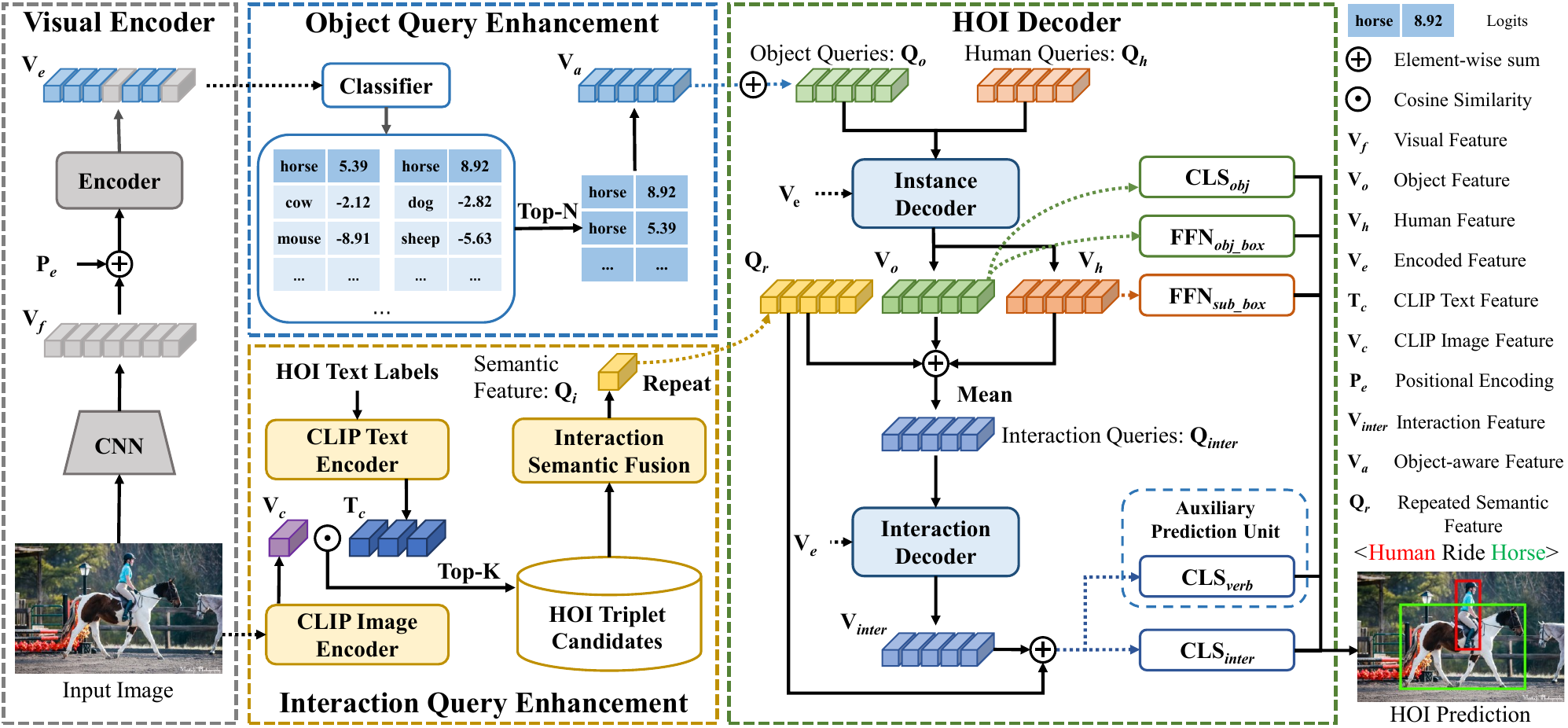}
\caption{Overall architecture of our DQEN. Compared with the previous works, our method contains two main components: Object Query Enhancement (OQE) and Interaction Query Enhancement (IQE). We propose OQE which selects the encoder feature with Top-\textit{N} confidence score to enhance object queries. Moreover, in IQE, we leverage CLIP's image-text matching to obtain HOI Triplet Candidates. The Interaction Semantic Fusion (ISF) module then derives semantic features from these labels to enhance interaction queries. Finally, we design an Auxiliary Prediction Unit (APU) to strengthen the model's ability to predict verbs, thereby improving its overall HOI prediction.}
\label{fig:fig2}
\end{figure*}

\section{Related Work}
\label{sec:related_work}
\subsection{HOI Detection}
Previous HOI detection methods can be categorized into two-stage and one-stage paradigms. The two-stage methods \cite{liu2022interactiveness,wu2022mining,zhang2021spatially,zhang2022exploring,zhou2020cascaded,li2020hoi} leverage a pre-trained object detector (e.g., DETR \cite{carion2020end}) to identify and localize all instances, encompassing both humans and objects. Subsequently, these methods systematically explore all possible human-object pairings, extract the features from the localized regions, and feed these cropped features into multi-path networks for processing. The multi-path networks typically consist of three main branches: a human branch, an object branch, and a pairwise interaction branch. To improve performance, various studies have proposed additional branches that integrate supplementary information. These enhancements encompass spatial features\cite{ulutan2020vsgnet}, word embeddings \cite{yuan2022detecting}, human pose information \cite{zhong2020polysemy}, or their combinations. Meanwhile, other research \cite{zhang2021spatially} has investigated the application of graph neural networks to model the relationships between human-object pairs.

However, two-stage methods suffer from expensive computation consumption due to their serial architecture for handling a large number of human-object pairs. To alleviate this issue, one-stage methods \cite{tamura2021qpic,liao2022gen,kim2021hotr,kim2023relational,xie2023category,ning2023hoiclip,zhang2021mining,zhong2022towards,cheng2022multi,yang2021rr} become popular in recent works. Recently, several HOI methods inspired by DETR \cite{carion2020end} have achieved promising performance. In particular, QPIC \cite{tamura2021qpic} pioneers the integration of DETR-based detection models into HOI detection, which efficiently harnesses comprehensive contextual cues across the image and accelerates the HOI learning process. GEN-VLKT \cite{liao2022gen} introduces a two-decoder pipeline that includes an instance decoder and an interaction decoder to enable a parallel forward process and employs distinct queries for humans and objects. Furthermore, HODN \cite{fang2023hodn} decomposes the instance decoder into a human decoder and an object decoder and is used to detect their corresponding targets independently. In the three-decoder pipeline, TED-Net \cite{wang2024ted} employs a dispersal attention mechanism on human and object decoding streams to capture interaction context, and an auxiliary discrimination mechanism to enhance HOI prediction accuracy. We adopt a two-decoder architecture and refine both branches, leveraging dual query enhancement to strengthen the model's ability to identify interactions, thereby improving its performance in detecting HOI compositions. Our work belongs to a one-stage end-to-end approach to study HOI detection.

\subsection{HOI Detection with External Knowledge}
There have been several approaches \cite{dong2022category,he2021exploiting,yuan2022rlip} that have utilized a wide variety of external knowledge to enhance the model’s ability to detect HOIs. CATN \cite{dong2022category} uses an external object detector to get the categories contained in the image and these categories would be transferred to corresponding word embeddings as final category priors. SG2HOI \cite{he2021exploiting} incorporates the scene graph information serving as the external knowledge to make up the visual cues. In addition, RLIP \cite{yuan2022rlip} leverages the VG dataset \cite{krishna2017visual} containing relational labels for training. Vision-Language Models (VLM) have shown strong generalization and have been adapted for HOI detection. GEN-VLTK \cite{liao2022gen} transfer the VLM to HOI detection through knowledge distillation \cite{hinton2015distilling,zheng2024restructuring}. HOICLIP \cite{ning2023hoiclip} uses the features obtained by the VLM visual encoder and proposes a new transfer strategy that uses visual semantic algorithms to represent verbs. In contrast to the aforementioned methods that directly apply external knowledge, our approach incorporates it in a more sophisticated manner, allowing for better identification of HOI categories.

\subsection{DETR and Its Variants}
Carion et al.\cite{carion2020end} proposed a Transformer-based end-to-end object detector named DETR (DEtection TRansformer) without using hand-designed components like anchor design and NMS. Despite its promising performance, the meaning of queries is unclear. Many subsequent studies \cite{zhu2020deformable,yao2021efficient,zhang2022dino} have investigated the connection between queries and spatial positions from various angles, seeking a deeper understanding of decoder queries in DETR. Deformable DETR \cite{zhu2020deformable} predicts 2D anchor points and designs a deformable attention module that focuses only on specific sampling points near a reference point. Efficient DETR \cite{yao2021efficient} enhances decoder queries by selecting the Top-\textit{K} positions from the dense predictions of the encoder. DINO \cite{zhang2022dino} enhances the queries with Top-\textit{K} selected encoder features. Our work builds on the standard DETR framework, with a primary emphasis on enhancing the effectiveness of decoder queries.

\section{Methods}
\label{sec:methods}

\subsection{Overall Architecture}
\label{subsec:overall_architecture}
The overall architecture of our model is presented in Fig. \ref{fig:fig2}. It includes a cascaded single-encoder plus double-decoder architecture, along with our proposed dual query enhancement modules. 
Given an input image \textit{I}, the primary visual features $\textbf{V}_{f}$ $\in$ $\mathbb{R}^{(H \times W) \times C^{'}}$ can be obtained through a CNN network followed by a flattening operation. It is supplemented with the positional encoding $\textbf{P}_{e}$ $\in$ $\mathbb{R}^{(H \times W) \times C^{'}}$ and then fed into the transformer encoder to generate the refined visual features $\textbf{V}_{e}$ $\in$ $\mathbb{R}^{(H \times W) \times C^{'}}$. $H$ and $W$ denote the height and width of the feature map obtained after downsampling the input image through CNN networks.

The HOI decoder part includes an instance decoder and an interaction decoder as marked by the green dashed box in Fig. \ref{fig:fig2}. Specifically, the instance decoder takes two sets of learnable embeddings as the inputs, i.e. the randomly initialized human queries $\textbf{Q}_h$ $\in$ $\mathbb{R}^{N_q \times C^{'}}$ and our enhanced object queries $\hat{\textbf{Q}}_{o} \in \mathbb{R}^{N_q \times C^{'}}$ (details presented in Section \ref{subsec:oos}), $N_q$ is the number of queries. Based on $\textbf{V}_{e}$, it returns two new visual features $\textbf{V}_{h}$ $\in$ $\mathbb{R}^{N_q \times C^{'}}$ and $\textbf{V}_{o}$ $\in$ $\mathbb{R}^{N_q \times C^{'}}$ for human and objects to predict their bounding boxes and category scores. Furthermore, a new interaction semantic feature $\textbf{Q}_{i} \in \mathbb{R}^{1 \times C^{'}}$ (details discussed in Section \ref{subsec:ise}) is repeated $N_q$ times and denoted as $\textbf{Q}_{r} \in \mathbb{R}^{N_q \times C^{'}}$. This repeated feature is then averaged with $\textbf{V}_h$ and $\textbf{V}_o$ to formulate the enhanced interaction queries $\textbf{Q}_{inter}$. It is then used as the input of the interaction decoder to extract interaction features $\textbf{V}_{inter} \in \mathbb{R}^{N_q \times C^{'}}$ from $\textbf{V}_{e}$.
Notably, a skip connection \cite{he2016deep} is applied between $\textbf{V}_{inter}$ and ${\textbf{Q}}_{r}$, which is used to predict the interaction category scores and verb category scores. 


\subsection{Object Query Enhancement}
\label{subsec:oos}





As aforementioned, the HOI triplets $<$\textit{humam, verb, object}$>$ is ultimately decoded from the visual feature $\textbf{V}_{e}$. 
As the output of a transformer encode, $\textbf{V}_{e}$ consists of $(H \times W)$ tokens of $C'$ channels. It can be viewed as an enriched version of its input (local visual features $\textbf{V}_{f}$) at each pixel, refined through the integration of global contextual information.
Naturally, those tokens on certain objects will contain clues about the human who interacts with them.
Therefore, our enhancement method aims to find the most representative object features that interacted with the human, in order to construct a better initialization for object queries.

\begin{figure}[htbp]
\centering
\includegraphics[width=\columnwidth]{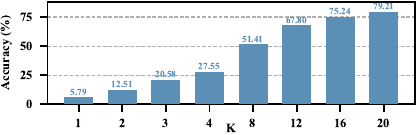}
\caption{Illustration of CLIP's image-text matching capability with the HOI accuracy of Top-\textit{K} matched texts in HICO-Det \cite{chao2018learning}. }
\label{fig:hoi_can}
\end{figure}

As illustrated within the blue dashed box in Fig. \ref{fig:fig2}, $\textbf{V}_{e}$ is fed into a linear layer to generate scores for each object category, which equals an object classification task. Those objects with the highest $N$ logit scores are good guesses for the following HOI detection. Thus, the corresponding features $\textbf{V}_{a}$ $\in$ $\mathbb{R}^{N \times C^{'}}$ are selected to enhance the randomly initialized object queries $\textbf{Q}_o$ $\in$ $\mathbb{R}^{N_q \times C^{'}}$ by an element-wise addition operation as follows,
\begin{equation}
    \hat{\textbf{Q}}_{o} = \textbf{Q}_{o} + \textbf{V}_{a} ,
\end{equation}
where $\hat{\textbf{Q}}_{o}$ is the enhanced object queries. To match with the dimension of $\textbf{Q}_o$, $N = N_q$. 

Noting that the gradients will propagate from the HOI losses back to this object classifier. It enables the model to effectively leverage high-quality object information to focus on objects involved in interactions. Meanwhile, $\textbf{V}_{e}$ is duplicated as input to the classifier, with gradient backpropagation disabled at this stage.
Otherwise, it may cause the model to overly focus on object information, thereby affecting the subsequent recognition of humans and interactions.

\subsection{Interaction Query Enhancement}
\label{subsec:ise}




Although finding the correct objects is important, identifying the interactions between humans and objects is the key that truly distinguishes HOI detection from other detection tasks. 
Therefore, the image-text matching capability of CLIP \cite{radford2021learning} is utilized to search most likely interactions, which are semantically fused to enhance interaction queries of the interaction decoder.

\textbf{HOI Triplet Candidates.}  
To align with the standard input format of the CLIP Text Encoder, HOI triplets $<$\textit{humam, verb, object}$>$ are converted into HOI Text Labels in the form of “A photo of a person [verb-ing] a/an [object]”. 
As presented in the {yellow dashed box} in Fig. \ref{fig:fig2}, the CLIP Image Encoder \textit{ImgEnc}$(*)$ is utilized to get the global visual feature $\textbf{V}_{c}$ $\in$ $\mathbb{R}^{D}$ from the input image \textit{I}, 
\begin{equation}
    \textbf{V}_{c} = 
    \textit{ImgEnc}(I) ,
\end{equation}
where $D$ is its embedding dimension. 
In addition, a text set $\mathcal{T}$ of HOI Text Labels is built as the input to the CLIP Text Encoder \textit{TextEnc}$(*)$. The text feature $\textbf{T}_{c} \in \mathbb{R}^{D \times N_{hoi}}$ can then be obtained as, 
\begin{equation}
    \textbf{T}_{c} = \textit{TextEnc}(\mathcal{T}) ,
\end{equation}
where $N_{hoi}$ is the number of the HOI categories.

\begin{figure}[htbp]
\centering
\includegraphics[width=\columnwidth]{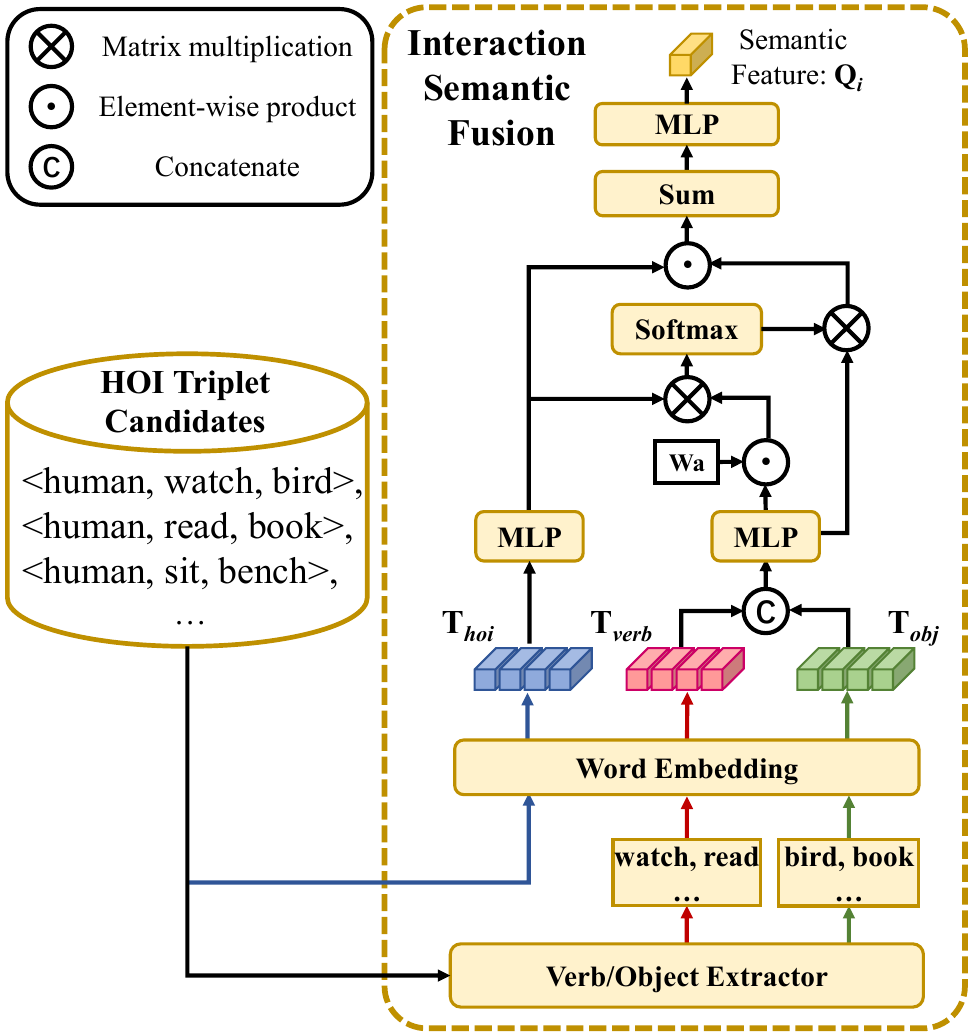}
\caption{The pipeline of the Interaction Semantic Fusion. }
\label{fig:ise}
\end{figure}

It is worth noting that the similarity $\textbf{m}_{sim}$ between $\textbf{V}_{c}$ and $\textbf{T}_{c}$ can be viewed as a coarse guess of the HOI within the image \textit{I}. As demonstrated in \ref{fig:hoi_can}, Top-8 matched text labels could cover over 50\% HOI ground-truth. Thus, it is intuitive to select the \textit{K} most relevant text labels to construct a set of HOI Triplet Candidates, denoted as $\mathcal{T}_{can}$. 
Such procedure can be formulated as follows,
\begin{equation}
    \textbf{m}_{sim} = \textbf{T}_{c}^{T}\textbf{V}_{c},
\end{equation}
\begin{equation}
    \mathcal{T}_{can} = \text{Top}K(\mathcal{T}, \textbf{m}_{sim}),
\end{equation}
where $\text{Top}K(*)$ denotes the function that returns the HOI triplets contained in the text label in $\mathcal{T}$ corresponding to the \textit{K} highest values in $\textbf{m}_{sim}$.




\textbf{Interaction Semantic Fusion.} Unlike the object features $\textbf{V}_{e}$ in Section \ref{subsec:oos}, the text features of $\mathcal{T}_{can}$ cannot be directly used to enhance the interaction queries, since they are mixed with the information of humans and objects. To refine the intertwined action and object information, we propose an Interaction Semantic Fusion module to extract the inherent interaction related features from decoupled texts.



As shown in Fig. \ref{fig:ise}, the words of $<$\textit{verb}$>$ and $<$\textit{object}$>$ are extracted from the HOI Triplet Candidates $\mathcal{T}_{can}$. Then, they are transformed to dense embeddings $\textbf{T}_{verb}$ $\in$ $\mathbb{R}^{K \times C_k}$ and $\textbf{T}_{obj}$ $\in$ $\mathbb{R}^{K \times C_k}$, respectively. Similarly, HOI Triplet $\mathcal{T}_{can}$ is embedded as $\textbf{T}_{hoi}$ $\in$ $\mathbb{R}^{K \times C_k}$. $C_k$ are 
the dimension of the word embedding.

To better capture semantic relationships, $\textbf{T}_{v}$ and $\textbf{T}_{o}$ are concatenated and fused to be a new embeddings $\textbf{T}_{vo}$ as,
\begin{equation}
     \textbf{T}_{vo} = \text{MLP}(Concat(\textbf{T}_{verb}, \textbf{T}_{obj})),
\end{equation}
while $\textbf{T}_{hoi}$ is also projected to the same semantic space as,
\begin{equation}
\hat{\textbf{T}}_{hoi} = \text{MLP}(\textbf{T}_{hoi}),
\end{equation}
where $\text{MLP}(*)$ is
a Multilayer Perceptron. For a given word embedding pair $\{\textbf{T}_{hoi}, \textbf{T}_{vo}\}$, their correlation matrix $\textbf{C}$ $\in$ $\mathbb{R}^{K \times K}$ is calculated as,
\begin{equation}
    \textbf{C} = (\textbf{w}_a \odot \textbf{T}_{vo}) \otimes \hat{\textbf{T}}_{hoi} ,
\end{equation}
where $\textbf{w}_{a}$ is a learnable parameters denoting the self-attention weights \cite{vaswani2017attention}, $\otimes$ and $\odot$ represent the matrix product and element-wise product, respectively. This correlation matrix $\textbf{C}$ is employed to re-weight $\textbf{T}_{vo}$ for a refined $\hat{\textbf{T}}_{vo}$ as,
\begin{equation}
    \hat{\textbf{T}}_{vo} = softmax(\textbf{C}) \cdot \textbf{T}_{vo},
\end{equation}
It is then fused with $\hat{\textbf{T}}_{hoi}$ to obtain the final \textbf{F} $\in$ $\mathbb{R}^{K \times C_k}$,
\begin{equation}
    \textbf{F} = \hat{\textbf{T}}_{vo} \odot \hat{\textbf{T}}_{hoi}.
\end{equation}

To integrate the interaction semantic features, a summation operation along the first dimension of \textbf{F} is applied, reducing it to a vector. Finally, a MLP is used to transform this vector into the interaction semantic feature $\textbf{Q}_{i}$ $\in$ $\mathbb{R}^{C^{'}}$,
\begin{equation}
    {\textbf{Q}_{i}} = \text{MLP}(\sum_{j=1}^{K} \textbf{F}(j, :)).
\end{equation}
It is then repeated and added with $\textbf{V}_h$ and $\textbf{V}_o$ to inject the semantic information into the interaction queries $\textbf{Q}_{inter}$.

\subsection{Auxiliary Prediction Unit}

Since the semantic features $\textbf{Q}_{i}$ inherently contain $<verb>$ information, it can be further used to predict categories of verbs to improve the model's representation of interactions. As illustrated in Fig. \ref{fig:fig2}, an additional verb classifier $\textbf{CLS}_{verb}$ is inserted to take the sum of $\textbf{Q}_{r}$ and $\textbf{V}_{inter}$ as the input. Its prediction $S_{verb}$ is then applied to modify the interaction prediction $S_{inter}$ from $\textbf{CLS}_{inter}$ as,
\begin{equation}
    S_{hoi} = S_{inter} + \alpha S_{verb} ,
\end{equation}
where $\alpha$ is a weighting parameter, $S_{hoi}$ is final HOI score.

\subsection{Training and Inference}
\label{subsec:trian_inference}

\textbf{Training.} During the training phase, we follow the previous works \cite{tamura2021qpic,zhang2021mining,liao2022gen,ning2023hoiclip}, and the Hungarian algorithm is used to assign the ground truth to the predictions. The loss of our model is composed of three parts, i.e. the bounding box regression loss $L_b$, the intersection-over-union (IOU) loss $L_u$, and the classification loss $L_c$, which is formulated as:

\begin{equation}
  \mathcal{L}_{\text {c}}= \sum_{i \in(h, o)} (\lambda_{b}\mathcal{L}_{b}^{(i)}+\lambda_{u}\mathcal{L}_{u}^{(i)})+\sum_{j \in(o, a)} \lambda_{c}^{(j)} \mathcal{L}_{c}^{(j)} ,
\end{equation}
where $\lambda_{b}$, $\lambda_{u}$ and $\lambda_{c}^{k}$ are the hyper-parameters for adjusting the weights of each loss. Superscripts (\textit{i}) and (\textit{j}) are the indices of predicted human/object bounding-boxes, and object/interaction categories, respectively. Moreover, We use cross-entropy (CE) loss to calculate the loss for the object classifier, denoted as $\mathcal{L_{\text{ce}}}$. So the final loss is given as, 
\begin{equation}
  \mathcal{L_{\text{total}}}= \mathcal{L}_{\text {cost }} + 
  \mathcal{L}_{\text {ce }} +
  \lambda_{kd} \mathcal{L}_{\text {kd}} , 
\end{equation}
in which $\mathcal{L}_{kd}$ is the knowledge distillation loss \cite{liao2022gen} of the CLIP model. Same as DETR \cite{carion2020end}, auxiliary losses are used on intermediate outputs of decoder layers.


\begin{table*}[htbp]
\caption{Comparison with state-of-the-art methods on HICO-Det. }
\centering
\renewcommand{\arraystretch}{1.1}
\begin{tabular}{lcccccccc}
\toprule
\multirow{3}{*}{Method} & \multirow{3}{*}{Backbone} & \multirow{3}{*}{Source} & \multicolumn{3}{c}{Default (mAP$\uparrow$)} & \multicolumn{3}{c}{Known Object (mAP$\uparrow$)} \\ 
 &  &  & Full & Rare & Non-Rare & Full & Rare & Non-Rare \\ 
\midrule
\textbf{\textit{Two-stage Methods:}}&  &  &  &  &  &  &  &  \\ 
SG2HOI\cite{he2021exploiting} & ResNet-50 & ICCV 2021 & 20.93 & 18.24 & 21.78 & 24.83 & 20.52 & 25.32 \\
SCG\cite{zhang2021spatially} & ResNet-50-FPN & ICCV 2021 & 29.26 & 24.61 & 30.65 & 32.87 & 27.89 & 34.35 \\
OCN\cite{yuan2022detecting} & ResNet-101 & AAAI 2022 & 31.43 & 25.80 & 33.11 & - & - & - \\ 
STIP\cite{zhang2022exploring} & ResNet-50 & CVPR 2022 & 32.22 & 28.15 & 33.43 & 35.29 & 31.43 & 36.45 \\ 
UPT\cite{zhang2022efficient} & ResNet-101-DC5 & CVPR 2022 & 32.62 & 28.62 & 33.81 & 36.08 & 31.41 & 37.47 \\ 
ViPLO\cite{zou2021end} & ViT-B/32 & CVPR 2023 & 34.95 & 33.83 & 35.28 & 38.15 & \textbf{36.77} & 38.56 \\
Open-Cat\cite{zheng2023open} & ResNet-101+ViT-B/16 & CVPR 2023 & 32.68 & 28.42 & 33.75 & - & - & - \\ 
ADA-CM\cite{lei2023efficient} & ResNet-50+ViT-B/16 & ICCV 2023 & 33.80 & 31.72 & 34.42 & - & - & - \\ 
PVIC\cite{zhang2023exploring} & ResNet-50 & ICCV 2023 & 34.69 & 32.14 & 35.45 & 38.14 & 35.38 & \underline{38.97} \\
CLIP4HOI\cite{mao2023clip4hoi} & ResNet-50+ViT-B/16 & NeurIPS 2023 & \underline{35.33} & \underline{33.95} & 35.74 & - & - & - \\
CMMP\cite{lei2024exploring} & ResNet-50+ViT-B/16 & ECCV 2024 & 32.26 & 33.53 & 33.24 & - & - & - \\
HOIGen\cite{guo2024unseen} & ResNet-50+ViT-B/16 & ACM MM 2024 & 34.84 & \textbf{34.52} & 34.94 & - & - & - \\
\midrule
\textbf{\textit{One-stage Methods:}}&  &  &  &  &  &  &  &  \\
HOI-Transformer \cite{zou2021end} & ResNet-50 & CVPR 2021 & 23.46 & 16.91 & 25.41 & 26.15 & 19.24 & 28.22 \\
QPIC \cite{tamura2021qpic} & ResNet-50 & CVPR 2021 & 29.07 & 21.85 & 31.23 & 31.68 & 24.14 & 33.93 \\
HOTR \cite{kim2021hotr} & ResNet-50 & CVPR 2021 & 25.10 & 17.34 & 27.42 & - & - & - \\
CDN \cite{zhang2021mining} & ResNet-50 & NeurIPS 2021 & 31.78 & 27.55 & 33.05 & 34.53 & 29.73 & 35.96 \\
CATN \cite{dong2022category} & ResNet-50 & CVPR 2022 & 31.86 & 25.15 & 33.84 & 34.44 & 27.69 & 36.45 \\
GEN-VLKT \cite{liao2022gen} & ResNet-50+ViT-B/32 & CVPR 2022 & 33.75 & 29.25 & 35.10 & 36.78 & 32.75 & 37.99 \\
ERNet \cite{lim2023ernet} & EfficientNetV2-S & TIP 2023 & 31.57 & 26.86 & 33.10 & 36.78 & 32.75 & 37.99 \\
Multi-Step \cite{zhou2023learning} & ResNet101 & ACM MM 2023 & 34.42 & 30.03 & 35.73 & 37.71 & 33.74 & 38.89 \\
HODN \cite{fang2023hodn} & ResNet-50 & TMM 2023 & 33.14 & 28.54 & 34.52 & 35.86 & 31.18 & 37.26 \\
MURN \cite{kim2023relational} & ResNet-50 & CVPR 2023 & 32.87 & 28.67 & 34.12 & 35.52 & 30.88 & 36.91 \\
HOICLIP \cite{ning2023hoiclip} & ResNet-50+ViT-B/32 & CVPR 2023 & 34.69 & 31.12 & 35.74 & 37.61 & 34.47 & 38.54 \\
TED-Net \cite{wang2024ted} & ResNet-50+ViT-B/32 & TCSVT 2024 & 34.00 & 29.88 & 35.24 & 37.13 & 33.63 & 38.18 \\
CEFA \cite{zhang2024plug} & ResNet-50+ViT-B/32 & ACM MM 2024  & 35.00 & 32.30 & 35.81 & \underline{38.23} & 35.62 & 39.02 \\
\hdashline
DQEN w/o TF & ResNet-50+ViT-B/32 & Ours & 35.23 & 32.79 & \underline{35.96} & \textbf{38.50} & \underline{36.42} & \textbf{39.12}  \\
DQEN w/ TF & ResNet-50+ViT-B/32 & Ours & \textbf{35.34} & 33.10 & \textbf{36.01} & 38.19 & 36.24 & 38.77 \\ 
\toprule
\multicolumn{9}{l}{* TF indicates the training-free approach.} \\
\end{tabular}
\label{tab: hico}
\end{table*}

\textbf{Inference.}
In order to leverage the zero-shot CLIP knowledge, we adopt a training-free approach during inference as,
\begin{equation}
    S_{tf} = \{ x \in \textbf{m}_{sim} \mid \text{Rank}(x) \leq R \} ),
\end{equation}
where $\text{Rank}(*)$ denotes the function that returns the normalized similarity score corresponding to the \textit{R} highest values in $\textbf{m}_{sim}$.
Following the protocol in \cite{liao2022gen}, the object score $S_o$ is obtained from the object classifier $\textbf{CLS}_{obj}$ after the instance decoder. Meanwhile, a set of human-object bounding-box pair ($B_{h}$,$B_{o}$) from $\textbf{FFN}_{sub}$ and $\textbf{FFN}_{obj}$ are generated. Combining with the HOI score $S_{hoi}$, the final HOI triplet score $S^{(n)}$ of \textit{n}-th HOI category is calculated as,
\begin{equation}
S^{(n)}=S_{hoi}^{(n)}+S_{o}^{(m)} \odot S_{o}^{(m)} + S_{tf}^{(n)},
\end{equation}
where the superscripts $(n)$ and $(m)$ indicate their corresponding category index.
Finally, the HOI triplets with the highest K confidence scores are processed by a triplet NMS as the final predictions. 


\section{Experiments}
\label{sec:experiments}

\begin{table}[htbp]
\caption{Comparison with state-of-the-art methods on V-COCO.}
\centering
\renewcommand{\arraystretch}{1.1}
\begin{tabular}{lclcc}
\toprule
Method & Backbone & $AP_{role}^{S_1}$ & $AP_{role}^{S_2}$ \\ 
\midrule
\textbf{\textit{Two-stage Methods:}} &  &  &  \\ 
SCG\cite{zhang2021spatially} & ResNet-50-FPN & 54.2 & 60.9 \\
OCN \cite{yuan2022detecting} & ResNet-50 & 64.2 & 66.3 \\
STIP \cite{zhang2022exploring} & ResNet-50 & \underline{65.1} & \textbf{69.7} \\ 
UPT \cite{zhang2022efficient}  & ResNet-50 & 59.0 & 64.5 \\ 
PVIC \cite{zhang2023exploring} & ResNet-50 & 62.8 & 67.8 \\
ViPLO\cite{zou2021end}  & ViT-B/32 & 60.9 & 66.6 \\ 
Open-Cat\cite{zheng2023open} & ResNet-101+ViT-B/16 & 61.9 & 63.2 \\ 
ADA-CM \cite{lei2023efficient} & ResNet-50+ViT-B/16 & 56.12 & 61.45 \\
CLIP4HOI \cite{mao2023clip4hoi} & ResNet-50+ViT-B/16 & - & 66.3 \\ 
CMMP \cite{lei2024exploring} & ResNet-50+ViT-B/16 & - & 61.2 \\
\midrule
\textbf{\textit{One-stage Methods:}}&  &  &  \\ 
HOI-Transformer \cite{zou2021end} & ResNet-50 & 52.9 & - \\
QPIC \cite{tamura2021qpic} & ResNet-50 & 58.8 & 61.0 \\
HOTR \cite{kim2021hotr} & ResNet-50 & 55.2 & 64.4 \\
CDN \cite{zhang2021mining} & ResNet-50 & 61.68 & 63.77 \\
CATN \cite{dong2022category} & ResNet-50 & 60.1 & - \\
Multi-Step \cite{zhou2023learning} & ResNet-50 & 61.3 & 67.0 \\
HODN \cite{fang2023hodn} & ResNet-50 & \textbf{67.0} & \underline{69.1} \\
HOICLIP \cite{ning2023hoiclip} & ResNet-50+ViT-B/32 & 63.5 & 64.8 \\
TED-Net \cite{wang2024ted} & ResNet-50+ViT-B/32 & 63.41 & 64.96 \\
CEFA \cite{zhang2024plug} & ResNet-50+ViT-B/32 & 63.53 & - \\ 
GEN-VLKT \cite{liao2022gen} & ResNet-50+ViT-B/32 & 62.41 & 64.46 \\
GEN-VLKT-OF \cite{liao2022gen} & ResNet-50+ViT-B/32 & 60.79 & 63.12 \\
\hdashline
Ours & ResNet-50+ViT-B/32 & 60.88 & 63.47 \\
\bottomrule
\multicolumn{4}{l}{* -OF indicates the optimal result provided by themselves in Github.} \\
\end{tabular}

\label{tab: V-COCO}
\end{table}

\subsection{Experimental Setting}

\textbf{Dataset.} We evaluate our model on two public benchmarks, HICO-Det \cite{chao2018learning} and V-COCO \cite{gupta2015visual}. HICO-Det consists of 47,776 images (38,118 for training and 9,658 for testing). It encompasses 600 HOI triplets, which are derived from a combination of 80 object categories and 117 action categories. Three evaluation types were designed based on the number of HOI categories in the training set, i.e. Full, Rare, and Non-Rare. Concretely, “Rare” includes 138 categories with fewer than 10 training samples, while the remaining 462 categories are classified as “Non-Rare”. V-COCO is a subset of the COCO dataset and comprises a total of 10, 396 images, with 5,400 allocated for training and 4,964 for testing. It features 29 distinct action categories, including 4 that represent body movements and do not involve interactions with objects. The dataset shares the same 80 object categories found in COCO, and from these, it forms a total of 263 unique HOI triplet classes. 

\textbf{Metrics.} We follow the protocol established in previous research \cite{chao2018learning,tamura2021qpic,zhang2021mining,liao2022gen} by employing mean Average Precision (mAP) as our performance metric. A HOI triplet prediction is deemed a true positive when it satisfies the following conditions: 1) The IoU between human and object bounding boxes and the ground truth exceeds 0.5. 2) The interaction category identified in the prediction is precise. For the HICO-Det dataset, we report performance in two settings: default setting and known object setting. In the former setting, the performance is evaluated on all test images while in the latter one, each AP is calculated on images that contain the target object class. For the V-COCO dataset, we report the role of mAP in two scenarios, where scenario 1 needs to predict the cases in which humans interact with no objects while scenario 2 ignores these cases.

\begin{table}[t]
    \caption{Zero-shot comparisons with SOTA methods.}
    \centering
    \begin{tabular}{lcccc}
    \toprule
    Method & Type & Unseen & Seen & Full \\
    \midrule
    VCL \cite{hou2020visual} & RF-UC & 10.06 & 24.28 & 21.43 \\
    ATL \cite{hou2021affordance} & RF-UC & 9.18 & 24.67 & 21.57  \\
    FCL \cite{hou2021detecting} & RF-UC & 13.16 &  24.23 & 22.01 \\
    GEN-VLKT \cite{liao2022gen} & RF-UC & 21.36 &   32.91 & 30.56 \\
    HOICLIP \cite{ning2023hoiclip} & RF-UC & 25.53 &   \textbf{34.85} & \textbf{32.99} \\
    Ours w/o TF & RF-UC & 25.09 & 33.19 & 31.57 \\
    Ours w/ TF & RF-UC & \textbf{26.79} & \underline{33.47} & \underline{32.13} \\
    \midrule
    VCL \cite{hou2020visual} & NF-UC & 16.22 &   18.52 & 18.06 \\
    ATL \cite{hou2021affordance} & NF-UC & 18.25 &   18.78 & 18.67 \\
    FCL \cite{hou2021detecting} & NF-UC & 18.66 &  19.55 & 19.37 \\
    GEN-VLKT \cite{liao2022gen} & NF-UC & 25.05 &  23.38 & 23.71 \\
    HOICLIP \cite{ning2023hoiclip} & NF-UC & 26.39 &  \textbf{28.10} & \textbf{27.75} \\
    Ours w/o TF & NF-UC & 26.05 & 24.71 & 24.97 \\
    Ours w/ TF & NF-UC & \textbf{26.67} & \underline{25.56} & \underline{25.78} \\
    \midrule
    ATL \cite{hou2021affordance} & UO & 5.05 & 14.69 & 13.08 \\
    FCL \cite{hou2021detecting} & UO & 0.00 & 13.71 & 11.43 \\
    GEN-VLKT \cite{liao2022gen} & UO & 10.51 & 28.92 & 25.63 \\
    HOICLIP \cite{ning2023hoiclip} & UO & \textbf{16.20} & \textbf{30.99} & \textbf{28.53} \\
    Ours w/o TF & UO & 11.59  & 29.08 & 26.17 \\
    Ours w/ TF & UO & \underline{15.70} & \underline{29.89} & \underline{27.53} \\
    \midrule
    GEN-VLKT \cite{liao2022gen} & UV & 20.96 & 30.23 & 28.74 \\
    HOICLIP \cite{ning2023hoiclip} & UV & 24.30 &  \textbf{32.19} & \textbf{31.09} \\
    Ours w/o TF & UV & \underline{24.49} & 30.27 & 29.46 \\
    Ours w/ TF & UV & \textbf{25.45} & \underline{30.84} & \underline{30.09} \\
    \bottomrule
    \end{tabular}
    
    \label{tab:zero-shot}
\end{table}

\textbf{Zero-shot Setting.} Following prior works \cite{liao2022gen,ning2023hoiclip}, we also conduct our zero-shot experiments in four different ways, namely Rare First Unseen Combination (RF-UC), Non-rare First Unseen Combination (NF-UC), Unseen Verb (UV), Unseen Object (UO). In the RF-UC setting, we select tail HOI categories as unseen categories, while in the NF-UC setting, we use head HOI categories as unseen categories. Under the UV and UO settings, some verb or object categories are not included in the training set, respectively. In the UC settings, all verb and object categories are present during training, but certain HOI combinations are omitted.

\textbf{Implementation Details.} For a fair comparison with previous methods \cite{liao2022gen,ning2023hoiclip,tamura2021qpic}, we use ResNet-50 as our backbone feature extractor, and the ViT-32/B CLIP variant for HOI Triplet Candidates Selection. The network is optimized using AdamW \cite{loshchilov2017decoupled} with a weight decay of $10^{-4}$.  The total number of training epochs is set to 90. We first train the model for 60 epochs with a learning rate of $10^{-4}$ that decreases by a factor of 10 for another 30 epochs. For the HOI Decoder, the layer number of the instance and interaction decoder are all set to 3. The number of HOI categories $N_{hoi}$ is 600 for HICO-Det and 263 for V-COCO. We set the number of queries $N_q$ to 64 and the number of channels $C^{'}$ to 256. Following GEN-VLKT \cite{liao2022gen}, we fine-tune the CLIP text embeddings initialized classifier with a small learning rate of $10^{-5}$ and set the loss weight $\lambda_{kd}$ to 20 during training. The weight of loss $\lambda_b$, $\lambda_u$, $\lambda_{c}^{o}$, $\lambda_{c}^{a}$ is set to 2.5, 1, 1 and 1, respectively. The embedding dimension of CLIP $D$ is 512, and the word embedding dimension $C_k$ is set to 512. We implement the zero-shot HOI experiments on HICO-Det. To enhance the extension to novel HOI categories, we freeze the CLIP initialized weights for interaction, verb and object classifiers. We restrict the HOI Text Labels to ‘seen’ categories during training, while we release this constraint to the ‘full’ 600 categories during inference. For the training-free approach, we take the value of \textit{R} to be 10. All experiments are conducted on 8 NVIDIA 4090 GPUs and the batch size is 16. The computational environment runs Ubuntu 22.04, with Python version 3.7, PyTorch version 1.10.0, torchvision version 0.11.0, and CUDA version 11.3.
\begin{table}[htbp]
\caption{Performance contribution of each component in DQEN on HICO-Det (Default). $\checkmark$ indicates that the module is used.}
\centering
\begin{tabular}{cccccc}
\toprule
APU & OQE & IQE & Full & Rare & Non-Rare \\ \midrule
- & - & - & 32.68 & 27.95 & 34.09 \\ 
$\checkmark$ & - & - & 33.72 & 30.02 & 34.83 \\  
-& $\checkmark$ & - & 33.28 & 29.10 & 34.53 \\
-& - & $\checkmark$ & 34.22 & 32.38 & 34.78 \\ \midrule
$\checkmark$ & $\checkmark$ & - & 33.50 & 29.47 & 34.71 \\
$\checkmark$ & - & $\checkmark$ & 34.47 & \textbf{32.83} & 34.96 \\
- & $\checkmark$ & $\checkmark$ & 34.26 & 30.36 & 35.43 \\ \midrule
$\checkmark$ & $\checkmark$ & $\checkmark$ & \textbf{35.23} & 32.79 & \textbf{35.96} \\ 
\bottomrule
\end{tabular}
\label{tab:ablation}
\end{table}


\begin{figure}[htbp]
\centering
\includegraphics[width=\columnwidth]{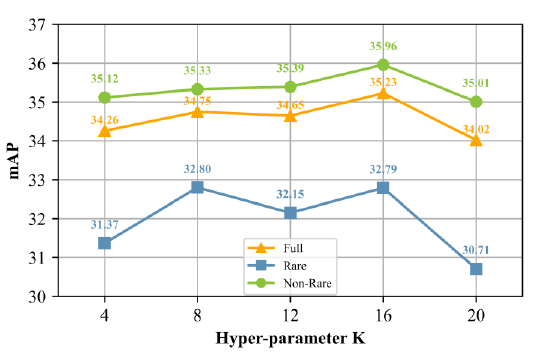}
\caption{Ablations Study of Hyper-parameter \textit{K} Selection. }
\label{fig:hyper_k}
\end{figure}

\begin{figure}[t]
\centering
\includegraphics[width=\columnwidth]{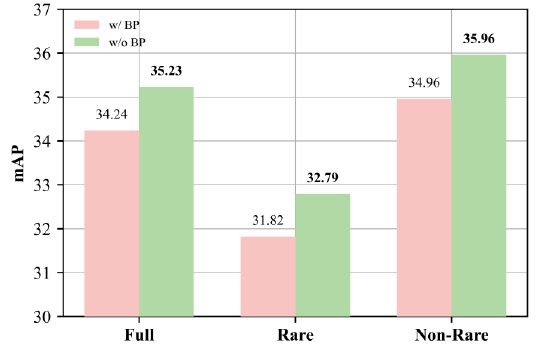}
\caption{Ablations Study of Gradient Truncation Design. }
\label{fig:ve_bp}
\end{figure}

\begin{table}[ht]
    \caption{Word Embedding Design. The Random embedding is following the Gaussian distribution.}
    \centering
    \begin{tabular}{lccc}
    \toprule
    Category & Full & Rare & Non-Rare \\
    \midrule
    Random & 33.46 & 29.99 & 34.49 \\
    CLIP Text Embedding & 34.66 & 32.55 & 35.29 \\
    Ours & \textbf{35.23} & \textbf{32.79} & \textbf{35.96} \\
    \bottomrule
    \label{tab:word_embedding}
    \end{tabular}
    
    \label{tab:word_embedding}
\end{table}

\subsection{Comparison to the State-of-The-Art}
\label{sec:comparison_sota}
\textbf{Standard Setting.} To verify the effectiveness of our proposed DQEN, we conducted experiments comparing it with existing state-of-the-art HOI detection methods. These comparisons were performed on both the HICO-Det and V-COCO datasets. Note that we use GEN-VLKT \cite{liao2022gen} as the baseline to remove Position-Guided Embedding, and change the human and object query initialization to the same value, which is done because the human and object queries are treated as a human-object pair under the same query subscript. For HICO-Det, as shown in Table \ref{tab: hico}, our DQEN outperforms the existing methods in both default and known object settings, which outperforms the baseline by \textbf{1.48} mAP, \textbf{3.54} mAP, and \textbf{0.86} mAP for Full, Rare and Non-Rare default setting, and \textbf{1.72} mAP, \textbf{3.67} mAP, and \textbf{1.13} mAP for Known Object setting, respectively. The queries in HOICLIP \cite{ning2023hoiclip} and TED-Net \cite{wang2024ted} share the similar initialization method of GEN-VLKT, our model outperforms both of these methods under the default and known object settings. In addition, our model improved the performance after adopting the training-free method, especially with \textbf{0.31} mAP improvement achieved under the default rare setting.

\begin{figure}[ht]
\centering
\includegraphics[width=\columnwidth]{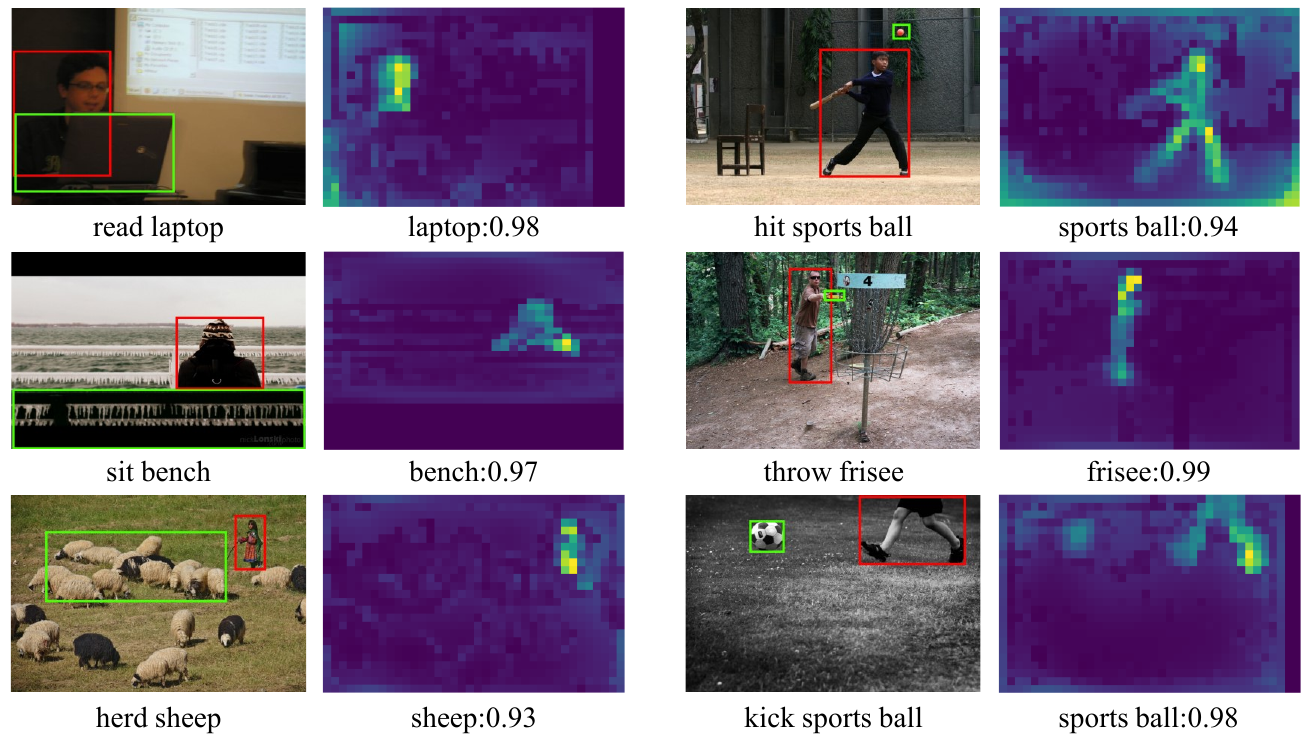}
\caption{Visualization of attention maps of encoder with the corresponding ground-truth bounding boxes. The corresponding predicted label and score are indicated below the attention map. Best viewed in color. }
\label{fig:oqe_atten}
\end{figure}

\begin{table}[ht]
\caption{Comparison of different methods in Object Detection based on mAP.}
\centering
\begin{tabular}{lc}
\toprule
Method & mAP ($\uparrow$) \\ \midrule
Baseline (GEN-VLKT) & \textbf{34.86} \\
Ours w/o OQE & 33.01 \\
Ours w/ OQE & 32.72 \\  \bottomrule
\label{tab:obj_map}
\end{tabular}

\label{tab:od_mAP_comparison}
\end{table}

Due to the higher average number of human-object pairs per image in the V-COCO dataset, as shown in Table \ref{tab: V-COCO}, the effectiveness of the OQE and IQE modules in extracting relevant object information and interactive semantic features is reduced compared to HICO-Det. As a result, performance improvements on the V-COCO dataset are less pronounced than those on HICO-Det. When compared to the results reproduced by our baseline \cite{liao2022gen}, we observe only a slight improvement. However, in both the S1 and S2 scenarios, our method outperforms ADA-CM and CMMP, which also utilize the CLIP model.

\textbf{Zero-shot Setting.} To verify the generalization of our method, we perform experiments in zero-shot settings on the HICO-Det dataset. The results are presented in Table \ref{tab:zero-shot}. Our method achieves competitive performance across multiple settings. In particular, compared with GEN-VLKT \cite{liao2022gen}, we achieve a \textbf{3.73} mAP increase under RF-UC settings across unseen categories and a \textbf{3.53} mAP improvement for unseen categories under the UV setting. These improvements show our model's strong generalization to unseen HOI categories. Furthermore, after adopting the training-free method, our model outperforms HOICLIP \cite{ning2023hoiclip} by \textbf{1.26} mAP in the RF-UC unseen setting, \textbf{0.28} mAP in the NF-UC unseen setting, and \textbf{1.15} mAP in the UV unseen setting.

\begin{figure}[ht]
\centering
\includegraphics[width=\columnwidth]{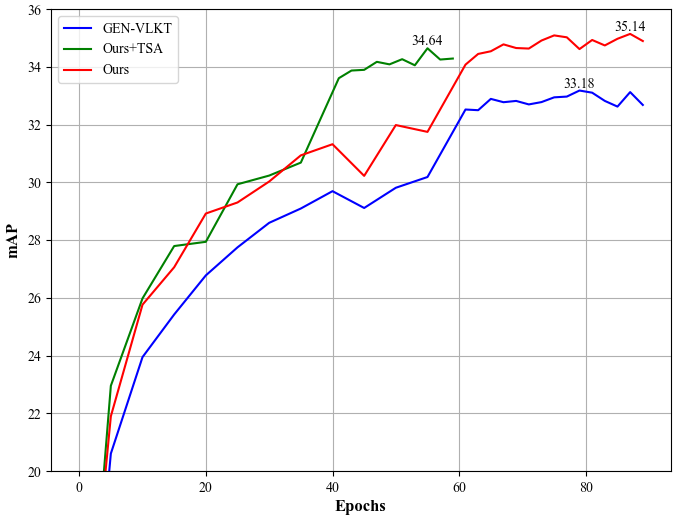}
\caption{Ours and GEN-VLKT's convergence curves. TSA denotes the training strategy adjustment is adopted to speed up convergence. GEN-VLKT‘s data is reproduced using the code available on GitHub. Experiments are conducted on the HICO-DET dataset. }
\label{fig:convergence}
\end{figure}

\begin{table}[ht]
\centering
\caption{Performance comparison on HICO-Det dataset and efficiency metrics.}
\begin{tabular}{lccccc}
\toprule
 & \multicolumn{3}{c}{Default (mAP$\uparrow$)} & \multicolumn{2}{c}{Efficiency} \\ 
{Method} & Full & Rare & Non-Rare & \#Params ($\downarrow$) & FPS ($\uparrow$) \\ \midrule
GEN-VLKT & 33.75 & 29.25 & 35.10 & \textbf{42M} & \textbf{31.01}  \\
TED-Net & 34.00 & 29.88 & 35.24 & 47M & 29.23 \\ 
Ours & \textbf{35.23} & \textbf{32.79} & \textbf{35.96} & 48M & 29.02 \\ \bottomrule
\end{tabular}
\label{tab:par_fps}
\end{table}

\begin{table}[t]
\centering
\caption{Performance comparison on HICO-Det dataset.}
\begin{tabular}{lcccc}
\toprule
 &  & \multicolumn{3}{c}{Default (mAP$\uparrow$)} \\ 
{Method} & Epoch ($\downarrow$) & Full & Rare & Non-Rare \\ \midrule
GEN-VLKT & 90 & 33.75 & 29.25 & 35.10  \\
HOICLIP & 90 & \underline{34.69} & 31.12 & \textbf{35.74}  \\
TED-Net & 120 & 34.00 & 29.88 & 35.24 \\ 
Ours w/o TF & \textbf{60} & 34.63 & \underline{31.79} & 35.49 \\ 
Ours w/ TF & - & \textbf{34.75} & \textbf{32.11} & \underline{35.54} \\ \bottomrule
\end{tabular}
\label{tab:comp_epoch}
\end{table}

\begin{figure*}[htbp]
\centering
\includegraphics[width=\linewidth]{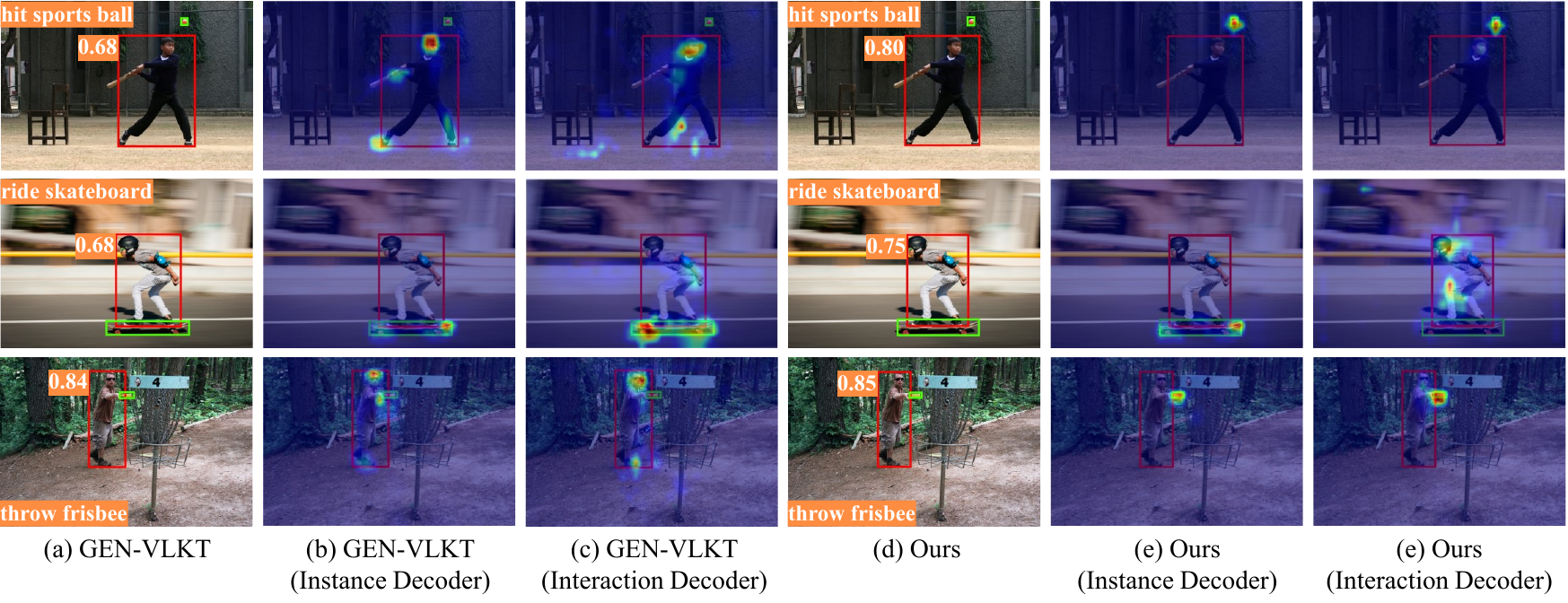}
\caption{Visualization of predictions. The orange rectangles containing scores indicate HOI scores, whereas those with words represent the predicted results. The columns indicate the detection result from GEN-VLKT \cite{liao2022gen} (a), the corresponding attention maps for object and interaction from instance decoder and interaction decoder (b, c), detection result from Ours (d), the corresponding attention maps for object and interaction from instance decoder and interaction decoder (e, f). Best viewed in color.}
\label{fig:encoder_featmap}
\end{figure*}

\subsection{Ablations Study}

\textbf{Network Architecture Design.}
To demonstrate the superiority of our designed framework, we conducted supplementary ablation experiments on the HICO-Det dataset. Firstly, we conduct experiments to illustrate the contribution of each component to the model. The results are shown in Table \ref{tab:ablation}, the data in the first row are the results reproduced after we made the modifications to GEN-VLKT (described in Section \ref{sec:comparison_sota}). Subsequently, we analyze the impact of the Auxiliary Prediction Unit (APU), Object Query Enhancement (OQE), and Interaction Query Enhancement (IQE) separately. It can be seen that each module is effective and the IQE module acts as the most influential factor in raising the full accuracy performance, resulting in a \textbf{1.54} mAP improvement over the baseline. It indicates that the introduction of interaction semantic information is helpful for HOI detection.

Moreover, we conduct a drop-one-out study, removing each proposed component individually to assess its unique contribution, this approach also allows us to validate the effectiveness of different module combinations. When three components lack IQE, the performance drops the most severely, it causes \textbf{1.48} mAP, \textbf{3.54} mAP, and \textbf{0.86} mAP drop for Full, Rare and Non-Rare default settings. The experimental results show that, among the three module combinations, only the APU and OQE combination performed worse than APU alone, with a performance drop of \textbf{0.22} mAP in Full setting, \textbf{0.55} mAP in Rare setting, and \textbf{0.12} mAP in Non-Rare setting. In contrast, the other combinations outperformed cases where each component was used individually. One possible reason for the performance degradation is that OQE and APU operate in different feature spaces, i.e. objects v.s. actions, which cannot effectively work together without the guidance of interaction features that connect them.




\textbf{Word Embedding Design.} We explore different initial strategies for word embedding in the Interaction Semantic Fusion module. Specifically, we conduct an extra experiment by using a set of random embeddings and CLIP text embedding to replace the original strategy respectively. As shown in Table \ref{tab:word_embedding}, our methods obtain a performance gain of \textbf{1.77} mAP and \textbf{0.57} mAP compared with the random and the CLIP Text Embedding under the full setting, respectively. It demonstrates that our word embedding design is more effective.

\textbf{Number of HOI Triplet Candidates.} In this part, we discuss the choice of hyper-parameter \textit{K} that determines the size of HOI triplet candidates $\mathcal{T}_{can}$. As presented in Fig. \ref{fig:hyper_k}, it can be seen that \textit{K} = 16 achieves the best performance in HICO-Det. For V-COCO, the Choice of Top-\textit{K} is set to 20 since the average number of variant human-object pairs per image is larger than HICO-Det. 

\textbf{Gradient Truncation Design.} As shown in Fig. \ref{fig:ve_bp}, allowing $\textbf{V}_{e}$ to participate in backpropagation (BP) directly results in a decrease in model accuracy, causing drops of \textbf{0.99}, \textbf{0.97}, and \textbf{1.00} in the Full, Rare, and Non-Rare settings, respectively. This precisely confirms our analysis in Section \ref{subsec:oos}, where allowing $\textbf{V}_{e}$ to participate in backpropagation within OQS impacts interaction recognition.

\textbf{Object Detection Test.}
In this part, we conduct experiments on the accuracy of $\textbf{CLS}_{obj}$ to validate OQE for object detection. As shown in Table \ref{tab:od_mAP_comparison}, we found that OQE does not improve the object detection performance of the model; instead, it slightly degrades it. However, experimental results demonstrate that it indeed benefits HOI detection. To further investigate the underlying reasons, we conducted a visualization of the selected encoder feature's attention map in Fig. \ref{fig:oqe_atten}. To see more clearly, we also draw the ground truth bounding boxes of humans and objects. Interestingly, we observe that, although the object categories produced by the classifier in OQE from the selected decoder features are correct, their corresponding attention maps highlight the person interacting with the object. For example, for the “person herding sheep”, the attention map highlights the person interacting with the sheep. Similarly, in the “person kicking a ball”, the model focuses more on the person's legs and feet. We believe that incorporating the features of the person interacting with the object into object queries would be more beneficial for subsequent human-object pair recognition.

\subsection{Model Complexity and Convergence}
Considering that the OQE, IQE, and APU introduce more parameters and more inference time, we conduct experiments about the number of parameters and FPS (Frames Per Second) on the HICO-Det test set. In terms of efficiency, our method has a slightly higher parameter count at 48M compared to GEN-VLKT's 42M and TED-Net's 47M. This increase in model complexity is justified by the improved performance across all categories. Despite the higher parameter count, our method still maintains a competitive FPS of 29.02, which is only slightly lower than GEN-VLKT's 31.01 and TED-Net's 29.23. This indicates that our method is capable of achieving competitive performance without significant sacrifices in computational efficiency.

Meanwhile, we also conducted experiments on the convergence of our method. As illustrated in Fig. \ref{fig:convergence}, it is important to note that the data presented in the figures are directly generated by the model without using the official evaluation protocol, which may result in slight discrepancies. With the help of our method, the model can converge faster. In addition, to further investigate the convergence properties of our proposed method, we modify the training strategy. Concretely, we changed the training epochs to 60 while keeping all other conditions constant, and also changed the epochs with reduced learning rates to the last 20 epochs. The results are shown in Table \ref{tab:comp_epoch}. The results indicate that our method can effectively help the model converge. To be specific, without the training-free method, comparing to our baseline GEN-VLKT, we can get \textbf{0.88}, \textbf{2.54} and \textbf{0.39} improvement in Full, Rare, and Non-Rare under Default setting on HICO-Det dataset with one-third reduction in the number of training epochs. The experimental results show that our method can enable the model to achieve better performance in fewer epochs. We envision that researchers, even those with limited GPU resources, can build upon our work to achieve further improvements.

\begin{figure*}[htbp]
\centering
\includegraphics[width=\linewidth]{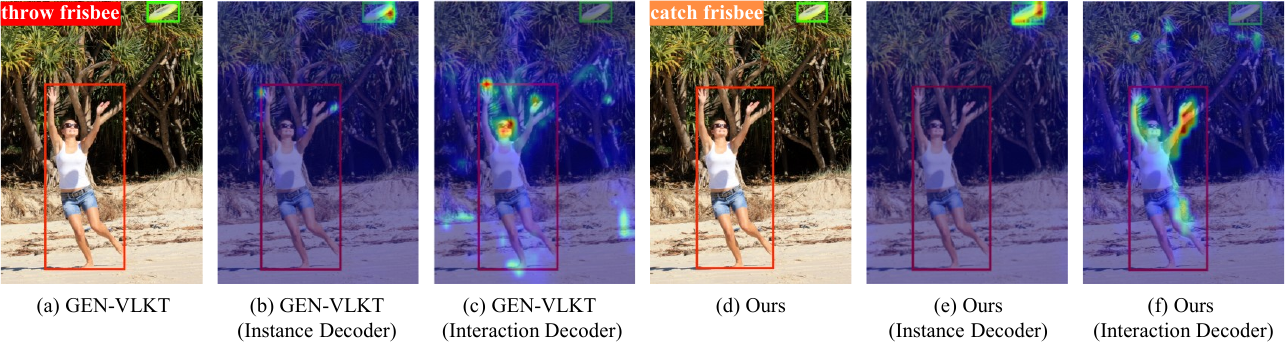}
\caption{Visualization of failed prediction in GEN-VLKT \cite{liao2022gen} and successfully predict in Ours. The red rectangle with words denotes incorrect predictions, whereas the orange rectangle with words denotes correct predictions. Best viewed in color.}
\label{fig:attn_map}
\end{figure*}

\subsection{Qualitative results}
The quantitative experimental results have demonstrated the effectiveness of our Dual Query Enhancement for HOI detection. As the qualitative comparison shown in Fig. \ref{fig:encoder_featmap}, although all the prediction results are correct, our model shows higher confidence in the correct outcomes. Furthermore, comparing the attention maps in columns (b) and (e), it can be observed that our model places greater emphasis on object information within interactions. Specifically, for the interaction of “throw a frisbee”, although our model has almost no difference in confidence compared to GEN-VLKT, on the attention maps of GEN-VLKT and Ours, it can be seen that our model focuses on the frisbee that is participating in the interaction, while GEN-VLKT focuses on the person. As shown in Fig. \ref{fig:attn_map}, GEN-VLKT mistakenly predicts the interaction “catch” as “throw”, whereas our method makes the correct prediction. Meanwhile, from the attention map of the interaction, it can be seen that our model pays more attention to the interaction area, i.e. human arms.

\section{Discussion}
Despite the effectiveness of the proposed method in certain scenarios, it exhibits marginal overall performance improvement, particularly on the V-COCO dataset. This performance degradation is primarily attributed to the higher average number of human-object pairs per image, which reduces the effectiveness of the OQE and IQE modules in capturing relevant object information and interactive semantic features. This limitation highlights the method's reduced capability in handling images with multiple co-occurring interactions, a scenario that closely resembles real-world HOI detection tasks. Addressing this challenge requires further investigation into strategies that enhance the model's robustness in complex interaction environments.

\section{Conclusion}
\label{sec:conclusion}

In this paper, we propose a DQEN, a one-stage method that explores dual query enhancement for DETR-based HOI Detection. For object query enhancement, our results demonstrate that object queries enhanced by the object-aware encoded features accurately convey more interaction information. For interaction query enhancement, enhancing interaction queries with semantic features can significantly improve the model's ability to represent and understand interaction features. In the future, we will further explore this work and seek more suitable query enhancement methods.

\bibliographystyle{IEEEtran}
\bibliography{main}

\section{Biography Section}
\vspace{-33pt}

\begin{IEEEbiography}[{\includegraphics[width=1in,height=1.25in,clip,keepaspectratio]{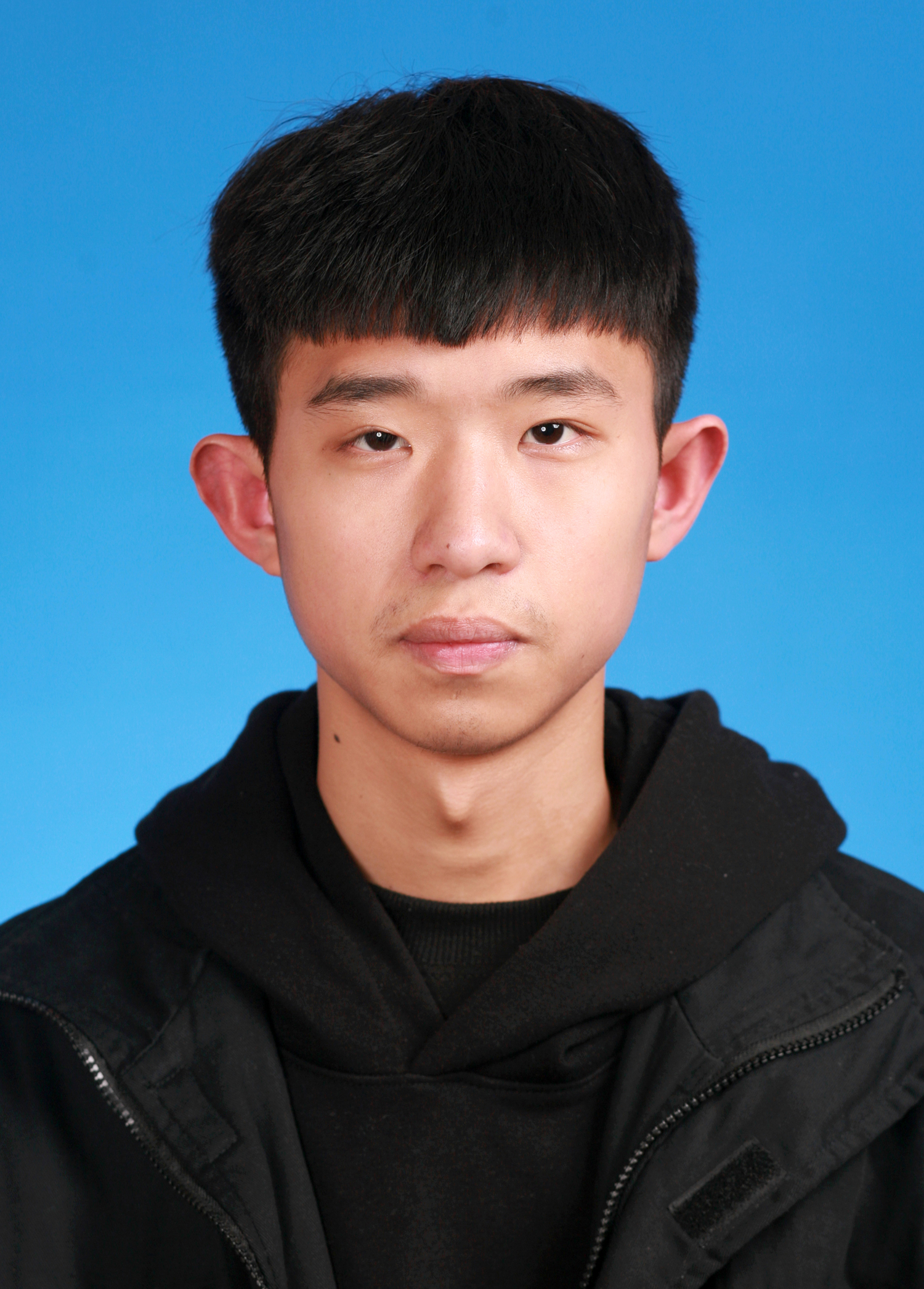}}]{Zhehao Li}
received his B.S. degree from Wenzhou University in 2023. He is currently pursuing a master's degree in the Faculty of Electrical Engineering and Computer Science, Ningbo University, Ningbo, China. His main research interests are in deep learning and human-object interaction detection.
\end{IEEEbiography}

\begin{IEEEbiography}[{\includegraphics[width=1in,height=1.25in,clip,keepaspectratio]{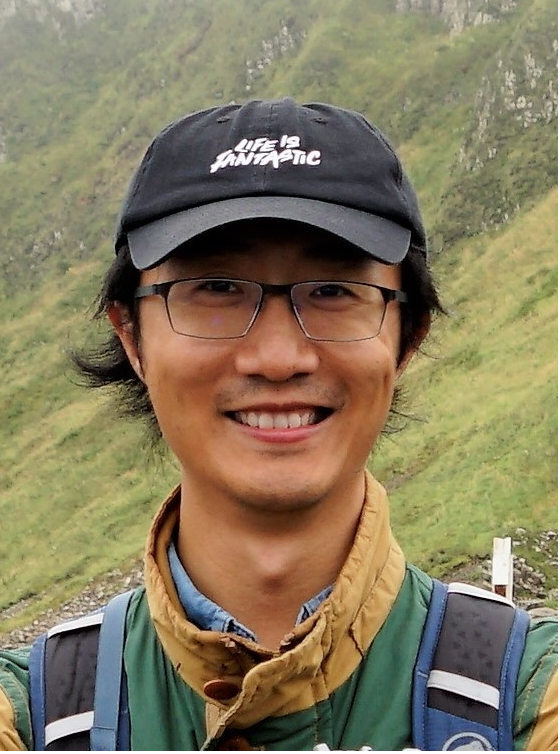}}]{Chong Wang}
received his Ph.D. degree from The University of Hong Kong in 2014. He was a research assistant and research associate in The University of Hong Kong and Hong Kong Polytechnic University, respectively. Since 2015, he has been with the Faculty of Electrical Engineering and Computer Science, Ningbo University, where he is currently an associate professor. His main research interests are in video anomaly detection, few-shot/zero-shot object detection, human-computer interaction, and neural network compression.
\end{IEEEbiography}

\begin{IEEEbiography}[{\includegraphics[width=1in,height=1.25in,clip,keepaspectratio]{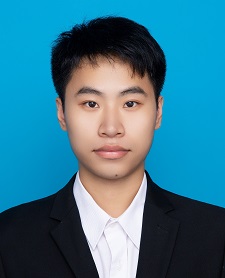}}]{Yi Chen}
received his B.S in Computer Science from Liren College of Yanshan University in 2022. He is currently a graduate student in the Computer Technology program at Ningbo University, located in Ningbo, China. His primary research interests include deep learning in computer vision and image generation.
\end{IEEEbiography}

\begin{IEEEbiography}[{\includegraphics[width=1in,height=1.25in,clip,keepaspectratio]{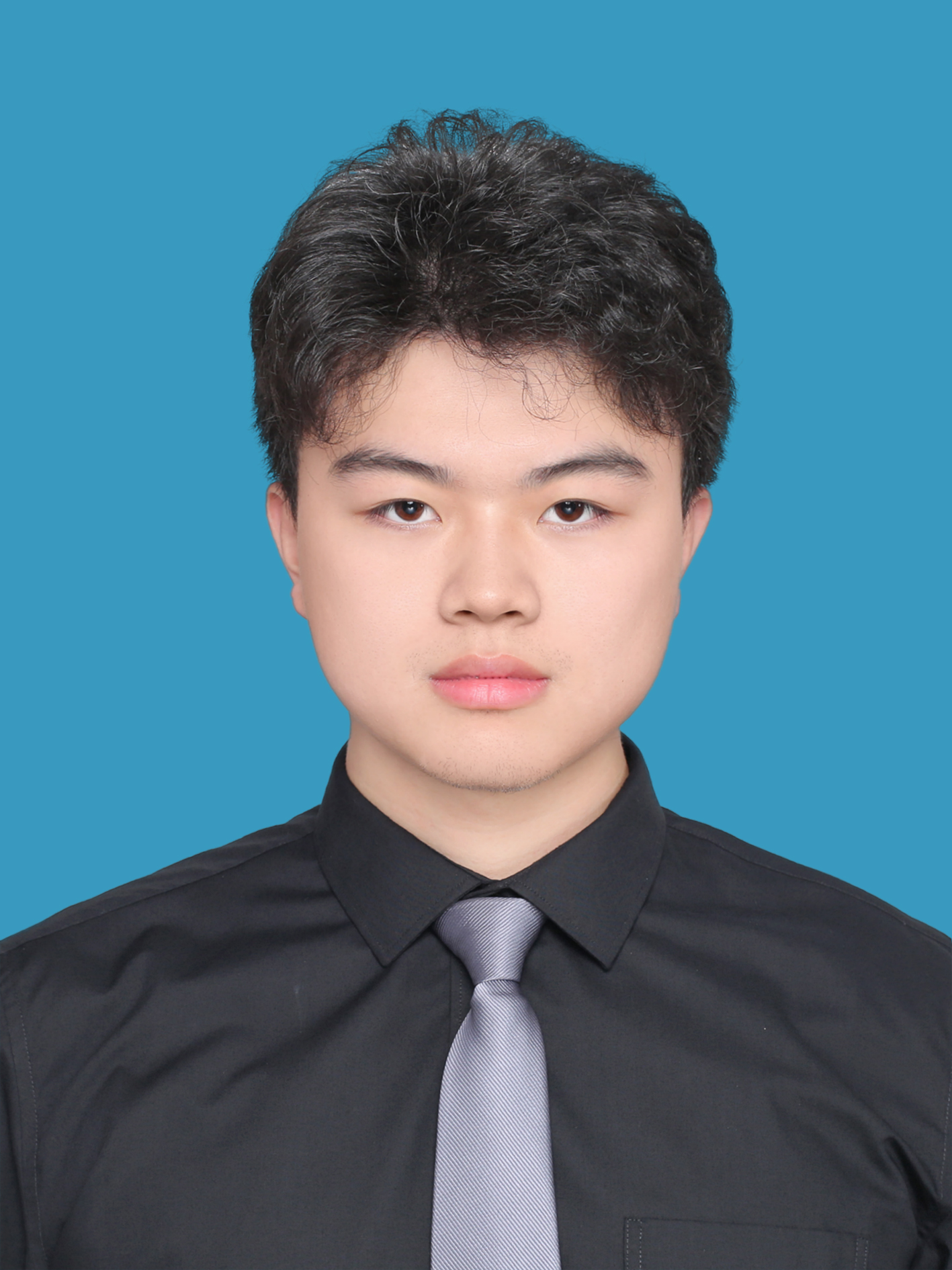}}]{Yinghao Lu}
received his B.S. degree from Suzhou Institute of Technology, Jiangsu University of Science and Technology in 2024. He is currently pursuing a master’s degree in the Faculty of Electrical Engineering and Computer Science at Ningbo University, Ningbo, China. His main research interests lie in the field of human-object interaction detection.
\end{IEEEbiography}

\begin{IEEEbiography}[{\includegraphics[width=1in,height=1.25in,clip,keepaspectratio]{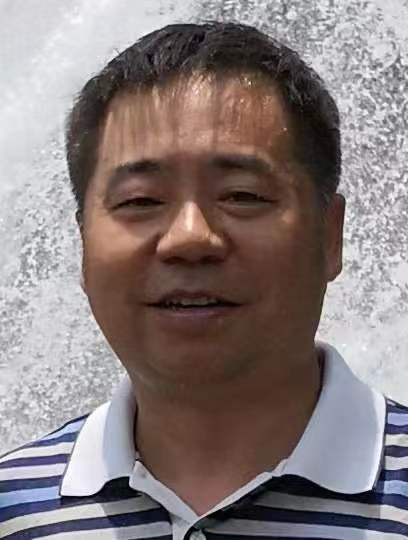}}]{Jiangbo Qian}
received the Ph.D. degree in computer science from Southeast University, China, in 2006. He was a Visiting Scholar with the Department of Computer and Information Science, The University of Michigan-Dearborn, USA. He is currently a Professor and Associate Dean with the Graduate School of Ningbo University,  China. He has published more than 150 papers in reputable international journals and conferences, including IEEE Transactions. His research interests include image processing, database management, streaming data processing, deep learning, computer vision, and hardware/software co-design.
\end{IEEEbiography}

\begin{IEEEbiography}[{\includegraphics[width=1in,height=1.25in,clip,keepaspectratio]{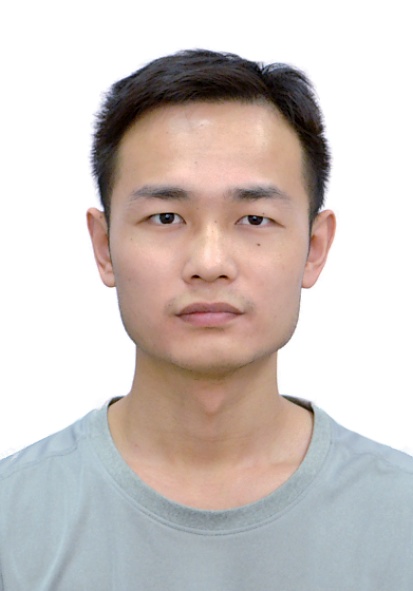}}]{Jiong Wang} received the B.E. degree from Shanghai University of Electric Power in 2016, followed by the M.E. and Ph.D. degrees from the Shenzhen University and Zhejiang University in 2019 and 2024, respectively. He is currently a lecturer in the Department of Computer Science at Ningbo University. His research interests focus on computer vision, multi-modal understanding, and vision-language model. 
\end{IEEEbiography}

\begin{IEEEbiography}[{\includegraphics[width=1in,height=1.25in,clip,keepaspectratio]{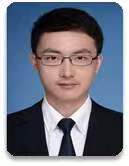}}]{Jiafei Wu}
received the B.S. degree from JXUFE in 2010, the M.S. degree and Ph.D. degree from the University of Hong Kong in 2012 and 2017, respectively. He is currently with the Zhejiang Lab. His research interests include deep learning, embedded system, and computational intelligence. 
\end{IEEEbiography}

\end{document}